\documentclass[sigconf]{acmart}

\usepackage{balance}
\usepackage{hyperref}
\usepackage{latexsym}
\usepackage{graphicx}
\graphicspath{{./images/}}
\usepackage{booktabs} 
\usepackage{color}  
\usepackage{amsmath}  
\usepackage{subcaption}
\usepackage{caption}
\usepackage{tikz}
\usepackage{colortbl} 
\usepackage{framed}
\usepackage{multirow}
\usepackage{multicol}
\usepackage{url}
\usepackage{verbatim}
\usepackage{cancel}
\usepackage{xspace} 
\usepackage[ruled,linesnumbered,vlined]{algorithm2e}
\usepackage{bbold} 
\usepackage{arydshln}

\newcommand{\utterance}[1]{\textit{#1}}

\newcommand{\myparagraph}[1]{\noindent \textbf{#1}.}

\newcommand{\praise}{\textsc{Praise}\xspace}
\newcommand{\convmix}{\textsc{ConvMix}\xspace}

\newcommand{\squishlist}{
	\begin{list}{$\bullet$}
		{ \setlength{\itemsep}{0pt}
			\setlength{\parsep}{3pt}
			\setlength{\topsep}{3pt}
			\setlength{\partopsep}{0pt}
			\setlength{\leftmargin}{1.5em}
			\setlength{\labelwidth}{1em}
			\setlength{\labelsep}{0.5em} } }
	\newcommand{\squishend}{
\end{list}  }

\AtBeginDocument{%
  \providecommand\BibTeX{{%
    \normalfont B\kern-0.5em{\scshape i\kern-0.25em b}\kern-0.8em\TeX}}}

\providecommand\BibTeX{{
 Bib\TeX}}

\acmConference[WWW Companion '25]{Companion Proceedings of the ACM Web Conference 2025}{April 28-May 2, 2025}{Sydney, NSW, Australia}
\acmBooktitle{Companion Proceedings of the ACM Web Conference 2025 (WWW Companion '25), April 28-May 2, 2025, Sydney, NSW, Australia}

\begin{document}

\fancyfoot{}
\pagenumbering{gobble}

\title{Preference-based Learning with Retrieval Augmented Generation for Conversational Question Answering}

	\author{Magdalena Kaiser}
	\affiliation{
		\institution{Max Planck Institute for Informatics
		\\ Saarland Informatics Campus}
        \country{Germany}}
	\email{mkaiser@mpi-inf.mpg.de}

	\author{Gerhard Weikum}
	\affiliation{
		\institution{Max Planck Institute for Informatics
		\\ Saarland Informatics Campus}
        \country{Germany}}
	\email{weikum@mpi-inf.mpg.de}

\renewcommand{\shortauthors}{Magdalena Kaiser \& Gerhard Weikum}


\begin{abstract}
Conversational Question Answering (ConvQA) involves multiple subtasks, 
i) to understand  incomplete questions in their context, 
ii) to retrieve relevant information, and iii) to generate answers.
This work presents \praise,
a pipeline-based approach for ConvQA that trains LLM adapters for each of the three subtasks.
As labeled training data for individual subtasks is unavailable in practice, 
\praise learns from its own generations using the final answering performance as feedback signal without human intervention
and treats intermediate information, like relevant evidence, as weakly labeled data.
We apply Direct Preference Optimization by contrasting successful and unsuccessful samples for each subtask.
In our experiments,  we show the effectiveness of this training paradigm: \praise shows improvements per subtask and achieves new state-of-the-art performance 
on a popular ConvQA benchmark, 
by gaining $15.5$ percentage points increase in precision over baselines.
\end{abstract}


\keywords{Conversational Question Answering, Retrieval Augmented Generation, Preference-based Learning}

\maketitle

\section{Introduction}
\label{sec:intro}

\begin{figure*} [ht]
	\centering
    \includegraphics[width=\textwidth]{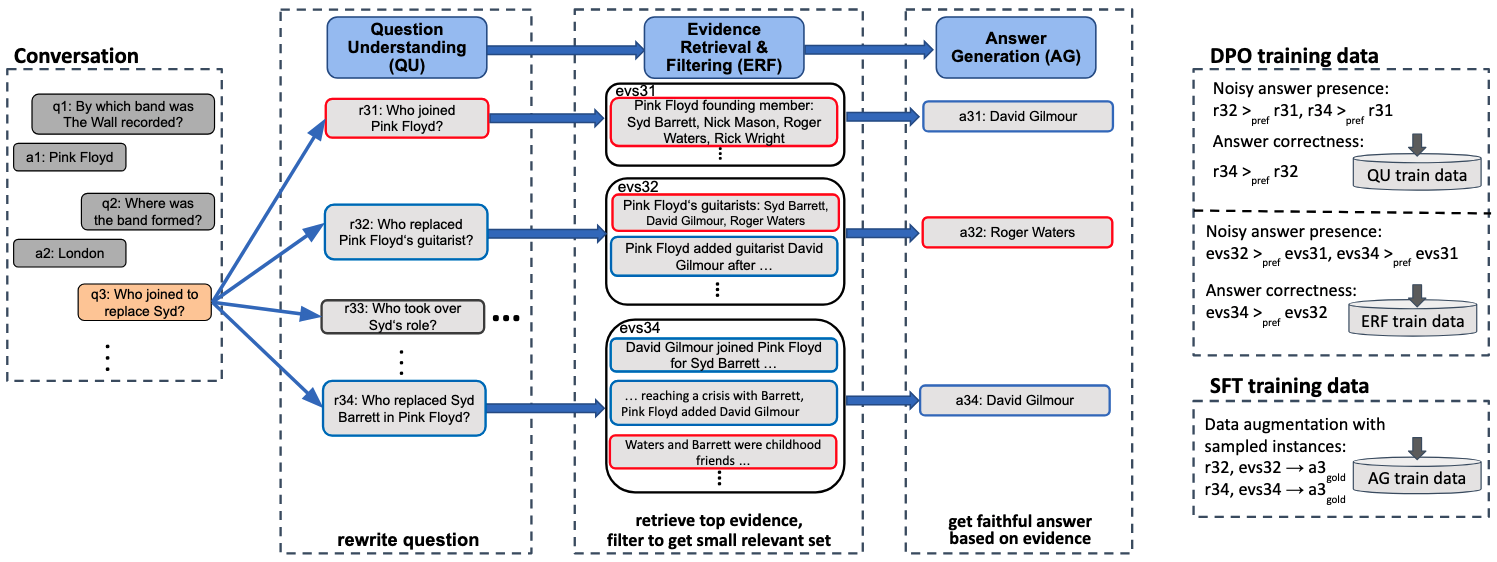}
	\caption{Overview and illustration of \praise (preferred/correct outputs are in blue, incorrect/uninformative outputs in red).}
	\label{fig:overiview}
\end{figure*}

\myparagraph{Motivation}
Conversational Question Answering (ConvQA) 
enables human-like interactions:
users pose a sequence of questions around a topic of interest, while the system keeps track of the conversation and is able to understand and answer incomplete, and even sloppily phrased, questions that make sense only in their conversational context.
An example is given on the left side of Fig.\ref{fig:overiview}, where the third turn's question $q_3: $ \textit{Who joined to replace Sid?} can be answered only by contextualizing it with the first question and its answer  $a_1: $ \textit{Pink Floyd}. 
Large Language models (LLMs) \cite{zhao2023survey}
have strong capabilities to contextualize the user's utterances, and are a key part of modern ConvQA systems.
However, LLMs are prone to hallucinations, giving non-factual answers or missing tangible provenance.
To mitigate these risks, LLM-based QA systems are usually coupled with
Retrieval Augmented Generation (RAG)
\cite{gao2023retrieval},
to provide the LLM with
retrieved evidence.
These considerations lead to a three-stage pipeline:
i) question understanding and contextualization,
ii) evidence retrieval and filtering/ranking,
iii) answer generation from best evidence.


This constitutes a bottleneck in training data.
In realistic settings, often only final answers are available as ground-truth,
but there is no ground-truth for question interpretation (like reformulated/completed questions) and no gold-standard evidence for the retrieval stage.
To overcome this problem and improve the performance of each subtask in the pipeline, 
we propose a sampling-based approach, by which we derive weakly-labeled intermediate training data, from observing final answers solely.

\vspace*{0.1cm}
\myparagraph{State-of-the-art Limitations}
Methods for ConvQA \cite{zamani2022conversational} contextualize questions with cues from prior turns, 
by reformulating questions, or 
by augmenting questions with selected pieces from the conversation.
All methods are driven by training data (or LLM fine-tuning data).
There are two main limitations: i) relying on intermediate supervision signals \cite{jeong2023phrase, mao2024chatretriever} - for evidence retrieval in RAG architectures, there is no cost-efficient way of compiling ground-truth data - and/or ii) relying on human feedback \cite{mo2023convgqr, kaiser2021reinforcement, vakulenko2021question}, which is costly to collect and often far from perfect. 
In contrast, \praise learns from final answering feedback in an automatic way.

\vspace*{0.1cm}
\myparagraph{Approach}
We introduce the \praise method (\textbf{P}reference-based Learning with \textbf{R}etrieval \textbf{A}ugmented \textbf{I}terative \textbf{SE}quence generation for ConvQA), a pipeline architecture, consisting of question understanding (QU), evidence retrieval and filtering (ERF) and answer generation (AG). 
Our QU model generates question reformulations so as to benefit the subsequent retrieval and answer generation.
The retrieval often yields long lists of potentially relevant evidence; feeding all these into the AG stage would incur high computational costs, if possible at all given the LLM's limited context length.
Therefore, the ERF stage of \praise contains 
a judiciously designed LLM-based evidence filtering technique,
efficiently operating with subsets of evidence pieces.
Finally, the AG stage is trained to give faithful answers
with tangible evidence containing the answer and question cues.
We perform parameter-efficient fine-tuning  of a Llama-3-8B-Instruct model \cite{touvron2023llama}. 
For each subtask, \praise samples generations from an initial model and learns from pairing successful and unsuccessful generations using Direct Preference Optimization (DPO)~\cite{rafailov2024direct}.
The intuition is that earlier tasks are optimized for later tasks via feedback signals without human intervention. 
%


\vspace*{0.1cm}
\myparagraph{Contribution}
This works makes the following contributions:
\begin{itemize}
    \item the \praise architecture for the ConvQA pipeline, where each stage can be optimized by learning with weakly-labeled intermediate data;
    \item a sampling-based technique for automatically deriving preference data
from the observed final answers and their ground-truth;
\item  experiments demonstrating the superior performance of \praise
on a popular ConvQA benchmark.
\end{itemize}
Our project website is available at \url{https://praise.mpi-inf.mpg.de}, our code and data is on GitHub: \url{https://github.com/magkai/PRAISE}.



\section{The \praise Method}
\label{sec:method}

Fig.~\ref{fig:overiview} gives an overview of the \praise architecture, along with an example conversation.
The following subsections explain the three stages of QU, ERF and AG, and
will refer to the example.




\subsection{Question Understanding (QU)}
This stage transforms the current user question $q_i$ 
into 
a small set of different,  more complete reformulations $r_{ij}$, by
considering question-answer pairs $q_\nu, a_\nu$ of the previous turns:
\begin{equation}
   r_{ij} \sim QU_{\praise}(q_i, \langle q_0, a_0, ..., q_{i-1}, a_{i-1} \rangle)
\end{equation}
\myparagraph{Initialization} We use an LLM in few-shot mode 
($QU_{few}$), 
showing it five examples on how to rewrite questions. The model is prompted to generate various surface forms for the same question. 

\vspace*{0.1cm}
\myparagraph{Training} 
For each original question, we obtain $x$ different reformulations ($x=5$ in our experiments); each of these is then fed into evidence retrieval and answer generation. 
The quality of the generated formulations is determined by the later stages as follows. A sample $r_{ij}$ is preferred if it fulfills the constraints:
\begin{enumerate}
    \item $r_{ij}$ results in retrieving evidence that mentions question entities and contains correct answers (answer presence);
    \item $r_{ij}$ leads to finding the right answer for this question (answer correctness).
\end{enumerate}
We create training data for fine-tuning the QU model based on these two constraints.
All question formulations that both constraints fulfill, are used for supervised fine-tuning (SFT), resulting in model $QU_{SFT}$.
Subsequently, we construct preference pairs for further training with direct preference optimization (DPO) 
($QU_{DPO}$):
\begin{itemize}
    \item $r_{ij} >_{pref} r_{ik}$ if $r_{ij}$ fulfills (1) but not $r_{ik}$ 
    \item $r_{ij} >_{pref} r_{il}$ if both fulfill (1) and only $r_{ij}$ fulfills (2) 
\end{itemize}
\vspace{0.1cm}
\noindent In Fig.~\ref{fig:overiview}, the rewritten questions $r_{32}$ and $r_{34}$ both fulfill (1) by retrieving relevant evidence, but only $r_{34}$ results in answering the question correctly and faithfully - fulfilling both (1) and (2).
Question $r_{31}$ results in a correct answer, but this answer is not faithful given the retrieved evidence (\textit{David Gilmour} is not mentioned in $evs31$), and therefore not considered as a positive sample.

\vspace*{0.1cm}
\myparagraph{Inference} The final $QU_{\praise}$ model uses greedy generation to obtain a completed question formulation.


\subsection{Evidence Retrieval and Filtering (ERF)}
This step identifies relevant pieces of evidence, tapping into knowledge graphs (KG), text corpora and web tables. 

\vspace{0.1cm}
\myparagraph{Initialization} The initial component for this step consists of the retrieval part only.
Our approach is agnostic to the underlying retrieval method: we make use of off-the-shelf tools, to obtain a large number of top-$n$ evidence ranked by BM25 scores.
The emphasis here is on high recall; obtaining high precision is left for the subsequent LLM-based filtering, as explained next.


\vspace*{0.1cm}
\myparagraph{Training} Each top-$n$ evidence is assigned a random id. 
We feed the rewritten question along with id-content pairs into the LLM for filtering. 
The objective of the filtering model is to output a small subset of top-$k$ ($k \ll n$) ids, $eids_{il}$, corresponding to the most relevant evidence $evs_{il}$ with $|evs_{il}| \leq k$ (so as to keep
the token costs as low as possible):
\begin{equation}
   eids_{il}  \sim ERF_{\praise}(r_{ij}, \langle eid_1: etext_1,..., eid_s: etext_s \rangle) 
\end{equation}

\noindent We start by training the filtering  model using SFT ($ERF_{SFT}$), by considering evidence that contains both question and answer entities as a positive sample.
In the example of Fig.\ref{fig:overiview}, 
the first and second evidence in set $evs_{32}$ and set $evs_{34}$, respectively, are considered as positive samples, containing the entities \textit{Pink Floyd}, \textit{Syd Barrett} and \textit{David Gilmour}. 
This is a weakly supervised technique, as we may face irrelevant snippets that merely contain the right entities by accident without being helpful for answering.
In Fig.\ref{fig:overiview}, the first evidence from $evs_{32}$ is such a spurious case.

For high precision, this initial ERF model is enhanced by preference learning as follows.
We sample evidence sets per question from $ERF_{SFT}$.
These pairs of question and evidence are passed to the AG stage to obtain answers.
Now we can exploit the available ground-truth for answers as a distant signal:
 evidence set $evs_{il}$ is preferred over set $evs_{im}$ ($evs_{il} >_{pref} evs_{im}$) if for the same question,  $evs_{il}$ in the input leads AG to the correct answer while  $evs_{im}$ does not.
In Fig.\ref{fig:overiview}, 
evidence set $evs_{34}$ is preferred over set $evs_{32}$, since $evs_{34}$ leads to the correct answer, but not $evs_{32}$.
We use the preference pairs as training data for DPO, resulting in an enhanced model $ERF_{DPO}$.

\vspace*{0.1cm}
\myparagraph{Inference} For tractability, we split the top-$n$ retrieval results (which could be $500$ or more) into smaller, manageable chunks of size $s$, and let the final $ERF_{\praise}$ model identify relevant evidence ids for each chunk. The union over the per-chunk outputs form the evidence that is given to the AG stage.
If less than $k$ evidence pieces are returned, we expand the set with more top-ranked evidence from the original BM25 scoring.

\subsection{Answer Generation (AG)}
This task generates answers $a_{im}$ for a question formulation $r_{ij}$, based on the filtered evidence content $\langle  etext_1,..., etext_k \rangle \in evs_{il}$:
\begin{equation}
    a_{im} \sim AG_{\praise}(r_{ij}, \langle  etext_1,..., etext_k \rangle)
\end{equation}
In the example of Fig.\ref{fig:overiview}, both $a_{31}$ and $a_{34}$ are correct answers, but only $a_{34}$ is faithful with respect to the provided evidence set.

\vspace*{0.075cm}
\myparagraph{Initialization} We initialize the AG model by fine-tuning an LLM with the available ground-truth answers from benchmark training data, resulting in $AG_{SFT}$. We use $QU_{few}$ and standard retrieval (without filtering) to create the input for the AG stage. 

\vspace*{0.1cm}
\myparagraph{Training} 
We improve the initial model with the sampled data from the previous stages (QU and ERF).
This way, the model learns from a larger variety of different surface forms and becomes more robust with respect to the order in which relevant and irrelevant information appears in the evidence set. 
%

%

\vspace*{0.1cm}
\myparagraph{Inference} We take the greedy model output along with multiple sampled outputs to  obtain an answer list for better comparison with other (ranking-based) methods.
\section{Experimental Setup}
\label{sec:setup}

\myparagraph{Benchmark}
We conduct our experiments on ConvMix\cite{christmann2022conversational}, a popular ConvQA benchmark over heterogeneous sources (from Wikidata and Wikipedia) with realistic questions over 5 different domains - books, movies, tv-series, music, soccer - coming from crowdworkers, with crisp sets of entities as answers. 

\myparagraph{Baselines}
We study the following methods:
\textbf{CONVINSE} \cite{christmann2022conversational}, a method that casts questions into intent-explicit structured representations and uses fusion-in-decoder to generate answers;
\textbf{EXPLAIGNN} \cite{christmann2023explainable}, which uses graph neural networks for reasoning over evidence and generating answers; 
\textbf{Mistral-7B + Graph + Memory} \cite{jain2024integrating}, which uses a graph structured representation with graph embeddings directly injected into an LLM (we compare with their best model);
\textbf{UniHGKR} \cite{min2024unihgkr}, with self-supervised pre-training, embedding alignment, and retriever fine-tuning  (we compare to its 7B model).
We also compare against end-to-end LLMs, specifically
\textbf{Mistral-7B-Instruct} \cite{jiang2023mistral}, in zero-shot mode and fine-tuned (reported numbers are taken from \cite{jain2024integrating});
\textbf{Llama-3-8B-Instruct} \cite{touvron2023llama}, in zero-shot and few-shot mode, and fine-tuned with  benchmark QA-pairs, with the entire conversation history in the input but without RAG.

\myparagraph{Metrics}
We  report Precision@1 (P@1), Mean Reciprocal Rank (MRR), and correct-answer-in-top-$5$ (Hit@5). 
Retrieval and
evidence filtering performance is measured by answer presence (AP) in the top-$k$ selected evidence ($k=500$ and $k=50$, respectively).

\myparagraph{Implementation Details}
We use Llama-3-8B-Instruct \cite{touvron2023llama} as underlying LLM for \praise and perform parameter-efficient fine-tuning 
by updating only ca. 12\% of its parameters per subtask.
For inference, we swap these trained adapters for the respective subtasks.\footnote{\footnotesize \url{https://huggingface.co/docs/peft/main/en/conceptual_guides/adapter}}
The initial pipeline for \praise consists of i) $QU_{init}$:  a Llama-3-8B-Instruct in few-shot mode  (with 5 examples and prompted to generate $5$ question formulations), ii)  $ERF_{init}$: evidence retrieval without filtering, (retrieving $n=500$ evidence snippets and taking the top $k=50$ evidence based on BM25 scores), and iii) $AG_{init}$: a Llama-3-8B-Instruct model fine-tuned  on benchmark QA-pairs for answer generation.
We adopt techniques from \cite{christmann2022conversational}, specifically, using the CLOCQ tool \cite{christmann2022beyond} for fact retrieval from the Wikidata KG and retrieving text snippets and table rows from Wikipedia.
We perform  multinomial beam-sampling, with beam size=$10$, for ERF training and AG inference.
The QU and ERF models are fine-tuned with SFT first, followed by DPO training for $1$ epoch each.
All experiments are run on a single GPU (NVIDIA Tesla A100, 80 GB HBM2e).
More details can be found in our code base (\url{https://github.com/magkai/PRAISE}).

\section{Results and Insights}

\subsection{Key findings}

\myparagraph{\praise achieves best performance}
Table \ref{tab:main-res} compares the performance of \praise to the baselines  on the \convmix test set. 
\praise outperforms all baselines by a substantial margin, gaining $15.5$ percentage points in P@1 over the second-best method (UniHGKR-7B - a model of similar size), with similarly strong improvements on the other metrics.
\praise also outperforms the end-to-end LLMs, showing the benefits of including RAG and a multi-stage pipeline.

\myparagraph{All pipeline stages contribute}
Table \ref{tab:praise-variants} shows the effects of different components in \praise, by comparing the full \praise model to its variants where one or two of  its components -- QU, ERF and AG -- are replaced with the respective initial models.
The variants with one \praise stage and two initial models show that each of the three components alone already improves performance. This trend is further enhanced with two components by \praise, 
and 
the full \praise outperforms all other configurations by a notable margin.

\label{sec:results}
\begin{table} [ht] 
	\newcolumntype{G}{>{\columncolor [gray] {0.90}}c}
	\resizebox{0.9\columnwidth}{!}{
	\begin{tabular}{l G c G}
		\toprule 
       \textbf{Method} $\downarrow$  \  \ \textbf{Metrics}  $\rightarrow$ &  \textbf{P@1} &	\textbf{Hit@5}	& 	\textbf{MRR}	\\ \toprule 
        CONVINSE \cite{christmann2022conversational} & $0.342$ & 	$0.386$ & $0.365$ 		 \\
         EXPLAIGNN \cite{christmann2023explainable}   & $0.406$  & $0.561$ &	$0.471$	 \\
    Mistral-7B + Graph + Memory \cite{jain2024integrating} & $0.445$ & $0.512$ & - \\
        UniHGKR-7B \cite{min2024unihgkr} & $0.465$ & $0.562$ & 	$0.514$ \\ \midrule
     Mistral-7B-Instruct (zero-shot) & $0.292$ & $0.346$ & - \\
     Mistral-7B-Instruct (fine-tuned) & $0.350$ & $0.400$ & - \\
    Llama-3-8B-Instruct (zero-shot)   & $0.403$	& $0.554$ &	$0.455$	 \\
      Llama-3-8B-Instruct (few-shot)   & $0.461$ &	$0.612$ &	$0.511$	 \\
       Llama-3-8B-Instruct (fine-tuned)   & $0.483$	& $0.640$ &	$0.536$	 \\ \midrule
        \praise  & $\mathbf{0.620}$ &  $\mathbf{0.746}$	& $\mathbf{0.665}$ \\
		 \bottomrule
	\end{tabular} }
	\caption{Main results comparing \praise to end-to-end LLMs and other competitors on the \convmix test set.}
	\label{tab:main-res}
    \vspace*{-0.2cm}
\end{table}
\begin{table} [ht] 
	\newcolumntype{G}{>{\columncolor [gray] {0.90}}c}
	\resizebox{\columnwidth}{!}{
	\begin{tabular}{l G c G c G}
		\toprule
       \textbf{Method} $\downarrow$ \ \ \textbf{Metrics}  $\rightarrow$ &  \textbf{P@1} & \textbf{Hit@5}	& 	\textbf{MRR}	& 	\textbf{AP@500}	& \textbf{AP@50}	\\ \toprule 
         $QU_{init} + ERF_{init} + AG_{init}$ &  $0.495$  &  $0.642$	& $0.543$ & $0.734$  & $0.486$\\ \midrule
         $QU_{\praise} + ERF_{init} + AG_{init}$  & $0.528$ &	$0.683$ &	$0.579$ & $\mathbf{0.767}$ & $0.536$\\
         $QU_{init} + ERF_{\praise} + AG_{init}$  & $0.559$ & $0.693$ &	$0.616$ & $0.734$ & $0.725$\\
        $QU_{init} + ERF_{init} + AG_{\praise}$  & $0.546$ & $0.678$ & $0.591$ & $0.734$  & $0.486$\\ \midrule
           $QU_{\praise} + ERF_{\praise} + AG_{init}$  & $0.582$ &	$0.712$	& $0.627$ & $\mathbf{0.767}$ & $\mathbf{0.758}$ \\
             $QU_{\praise} + ERF_{init} + AG_{\praise}$  & $0.580$ & $0.708$ &	$0.624$ & $\mathbf{0.767}$ & $0.536$  \\
               $QU_{init} + ERF_{\praise} + AG_{\praise}$  & $0.587$ & $0.723$	& $0.635$ &  $0.734$ & $0.725$\\ \midrule
     $\praise full (QU+ERF+AG)$ &  $\mathbf{0.620}$ &  $\mathbf{0.746}$	& $\mathbf{0.665}$ & $\mathbf{0.767}$ & $\mathbf{0.758}$\\  
		 \bottomrule
	\end{tabular} }
	\caption{Effect of different pipeline components in \praise.}
	\label{tab:praise-variants}
    \vspace*{-0.2cm}
\end{table}

\begin{table} [ht] 
	\newcolumntype{G}{>{\columncolor [gray] {0.90}}c}
	\resizebox{\columnwidth}{!}{
	\begin{tabular}{l G G c c G G c c G G} 
		\toprule
       \textbf{Domain $\rightarrow$}   &     \multicolumn{2}{G}{\textbf{Books}} & \multicolumn{2}{c}{\textbf{Movies}} & \multicolumn{2}{G}{\textbf{TV series}}    & \multicolumn{2}{c}{\textbf{Music}}    & \multicolumn{2}{G}{\textbf{Soccer}} \\ 
         \textbf{Method $\downarrow$} &  \textbf{P@1} & \textbf{AP@50} &  \textbf{P@1} & \textbf{AP@50} &  \textbf{P@1} & \textbf{AP@50} &  \textbf{P@1} & \textbf{AP@50} &  \textbf{P@1} & \textbf{AP@50} 
       \\
       \toprule 
     $\praise \ init$ &  $0.491$ &	$0.477$ &	$0.530$ &	$0.438$ &	$0.506$ &	$0.500$ &	$0.505$ &	$0.510$ &	$0.443$	& $0.510$  \\
    $\praise \ full$ & $\mathbf{0.650}$ &	$\mathbf{0.814}$ &	$\mathbf{0.634}$	& $\mathbf{0.721}$ &	$\mathbf{0.614}$ &	$\mathbf{0.716}$ &	$\mathbf{0.630}$ &	$\mathbf{0.779}$ & 	$\mathbf{0.572}$ &	$\mathbf{0.760}$ \\
		 \bottomrule
	\end{tabular} }
	\caption{Domain-wise results for \praise compared to the initial pipeline ($\mathbf{QU_{init} + ERF_{init} + AG_{init}}$).} 
	\label{tab:domain-res}
     \vspace*{-0.2cm}
\end{table}



\begin{table} [ht] 
		\begin{tabular}{p{8cm}}
			\toprule
   \textbf{Conversation:} \utterance{What number of Pulitzers has John Updike won? 2; What book won the author the award first? Rabbit Is Rich; name of the other? Rabbit at Rest; publication year of the book?}   \\ \hdashline
             \textbf{QU (rewritten question):} \utterance{What is the publication year of the book ``Rabbit at Rest'' by John Updike?} \\
           \textbf{ERF  (filtered evidence):}   [\textcolor{red}{\textit{id-150}}: John Updike, award received, National Book Critics Circle Award, point in time, 1990, for work, Rabbit at Rest], [\textit{id-391}: Rabbit at Rest, Publication date, 1990], [\textit{id-127}: Rabbit at Rest is a 1990 novel by John Updike] ...    \\
    	   \textbf{AG  (generated answer): }  \utterance{1990} \\ \midrule
	 \textbf{Conversation:} \utterance{What was Jack Nicholson's last film of the 1990s? As Good as It Gets; Who directed?}   \\ \hdashline
             \textbf{QU (rewritten question):} \utterance{Who directed the film``As Good as It Gets'' starring Jack Nicholson?} \\
           \textbf{ERF  (filtered evidence):}   [\textit{id-389}: As Good as It Gets is a 1997 American romantic comedy film directed by James L. Brooks from a screenplay he co-wrote with Mark Andrus.], [\textcolor{red}{\textit{id-307}}: James L. Brooks and Jack Nicholson, doing what they do best, combine smart dialogue and flawless acting...],  [\textit{id-187}: As Good as It Gets, director, James L. Brooks] ...    \\
    	    \textbf{AG  (generated answer): }  \utterance{James L. Brooks} \\ 
           \midrule
            \textbf{Conversation:} \utterance{ On which date was the soccer player Neymar born? 5 February 1992; Complete name? Neymar da Silva Santos Júnior; How tall? 5 ft 9 in; Position at which he plays?}   \\ \hdashline
             \textbf{QU (rewritten question):} \utterance{What position does Neymar play in soccer?} \\
           \textbf{ERF  (filtered evidence):}   [\textit{id-231}: ... Neymar was voted the Best player and Best Forward], [\textit{id-296}: Neymar, Personal information, Position(s), Left winger , attacking midfielder , forward],  [\textit{id-206}: List of international goals scored by Neymar, Neymar is a Brazilian professional footballer who plays as a forward.]    \\
    	   \textbf{AG  (generated answer): }  \utterance{Left winger, forward} \\
                     \bottomrule
	\end{tabular}  
	\caption{Examples of outputs generated by the three stages in \praise  (three evidence pieces are shown due to space limitations, irrelevant evidence indicated in red).} 
	\label{tab:praise-examples}
    \vspace*{-0.6cm}
\end{table}

\subsection{In-depth analysis}
\myparagraph{\praise achieves high answer presence}
Table \ref{tab:praise-variants} gives insights on answer presence.
In the retrieval step, top-$500$ evidence snippets are obtained,  resulting in an answer presence of ca. $73\%$. 
When using the reformulated questions by $QU_{\praise}$, AP is increased by $3\%$.
For filtering, when we simply use BM25 scores to obtain top-$50$ evidence,
the AP drops sharply: down to $49\%$ for the initial model. 
In contrast, the ERF of \praise retains high answer presence at $75.8\%$, 
when reducing the initially retrieved top-$500$ down to $50$ evidence pieces.
%
%
%
%
Another interesting observation is that the intermediate $ERF_{SFT}$ model has a much lower answer presence (ca. $51\%$)
than our $ERF_{\praise}$ model.
This underlines that the DPO-based fine-tuning with feedback from the AG model is crucial 
to optimize the preceding ERF stage. 
Overall, this reconfirms our intuition that optimizing the subtasks of the pipeline, with training samples automatically
derived from question-answer pairs, is beneficial.

\myparagraph{Preference optimization is effective}
Instead of performing preference learning, we could also directly use the generations that successfully lead to the correct answer for further supervised learning (replacing both, $QU_\praise$ and $ERF_\praise$, with $QU_{SFT}$ and $ERF_{SFT}$ respectively, in the pipeline).
This results in a performance decrease ($P@1=0.585$ instead of $0.620$ for \praise), demonstrating that using DPO after an initial supervised fine-tuning step is more effective.



\myparagraph{\praise creates fluent reformulations} 
While the original questions are relatively short -- ca. $5.6$ words on average, the reformulations generated by $QU_{init}$  have an average length of $9.3$ words, and the ones by  $QU_{\praise}$ go up to $10.1$. Our method successfully contextualizes the questions with cues from previous turns in the conversation, while keeping the questions natural and fluent.
This is confirmed by analyzing $200$ reformulations manually.

Table~\ref{tab:praise-examples} shows the outputs generated by the three stages in \praise, for some exemplary conversations from \convmix. 


\myparagraph{Performance improvements across domains}
Table~\ref{tab:domain-res} shows  domain-wise results for \praise, in comparison to the initial pipeline ($QU_{init} + ERF_{init} + AG_{init}$) on the \convmix testset.
The performance of \praise is robust across domains, consistently improving upon the initial pipeline
(with up to  $16$ percentage points increase in $P@1$ and $34$ percentage points increase in $AP@50$).
The answer presence of \praise is highest for books ($81\%$) and lowest for movies and  tv-series ($72\%$). 
Nonetheless, \praise achieves high precision for all of these three domains (in the range of $61$ to $65 \%$). 
We attribute this to the rich signals from the underlying LLM's parametric memory (from its original pre-training). 
Thus, \praise incorporates informative RAG without losing the benefits from LLM memory.

\myparagraph{Runtime analysis and memory consumption}
\praise requires around $5.4$ s on average to answer a conversational question.
This time is spent in the respective stages as follows:
$QU_\praise$ takes around $0.5$ s,  $ERF_\praise$ takes $4.6$ s (with $1.3$ s spent for evidence retrieval and $3.3$ s for multi-step evidence filtering), and 
$AG_\praise$ requires $0.34$ s.
The time for $ERF_\praise$ can be decreased further by either retrieving less evidence (currently $500$) or splitting into larger chunks for evidence filtering  (currently, we split into $10$ chunks with $50$ evidence pieces each, resulting in $10$ calls to the filtering model). 
Instead of loading three LLMs (one for each subtask) into memory, 
we switch our trained task-specific adapters, so that we only need to keep one LLM in memory. This way, \praise requires around $34$ GB of GPU memory.
\section{Related Work}
\label{sec:related}

\myparagraph{Conversational Question Answering}
There are two paradigms to handle incomplete conversational questions, i) question rewriting - utilizing information from the conversation history to cast the question into a self-contained form, and ii) question contextualization - augmenting the question with relevant information from previous turns (see survey \cite{zamani2022conversational}).
We opt for rewriting since we use LLMs in our pipeline, 
which benefit from fluent input. 
Many earlier works required human rewrites (e.g., \cite{vakulenko2021question}), which is costly to collect and not necessarily optimal for retrieval.
Recent works aim to improve rewriting for retrieval (e.g., \cite{mo2023convgqr, zhang2024adaptive, yoon2024ask}), whereas \praise optimizes for retrieval as well as answering performance. 

\myparagraph{RAG for ConvQA}
The RAG paradigm has recently received great attention as an enhancement of LLM-based generation including question answering to mitigate the risk of hallucination, and to make QA (more) faithful \cite{guu2020retrieval,lewis2020retrieval,gao2023retrieval}. 
Many ConvQA methods make use of LLMs, dense or sparse neural retrieval and various kinds of representation learning (e.g., \cite{christmann2023explainable,hai2023cosplade, jain2024integrating, jeong2023phrase, mao2024chatretriever, mo2023convgqr,tran2024conversational}). 

\myparagraph{Learning from Feedback for QA}
Reinforcement Learning has been used as means to improve QA models with answering performance as feedback (e.g., \cite{kaiser2024robust, yang2023prca}).
In the context of LLMs, Reinforcement Learning from Human Feedback (RLHF) \cite{ouyang2022training} has been proposed to align LLMs to human preferences. 
Direct Preference Optimization (DPO) \cite{rafailov2024direct}, and variants, were introduced as less complex and more stable options.
Since collecting human preference is costly, an alternative is to use automatic feedback (e.g., \cite{yuan2024self, peng2023check, zhang2024adaptive}). 
\section{Conclusion}

This work presents \praise, an LLM-based pipeline for ConvQA, where earlier tasks are improved with feedback from later tasks.
To achieve these enhancements, we use DPO with training data derived from contrastive pairs of successful and unsuccessful samples. 
This method is particularly suitable for domains in which the task can be divided into subtasks 
and when there is task-level feedback, but only sparse intermediate annotations, available. 
One direction for future work is to apply the \praise methodology to create agents for such domains, like finance or customer service. 




\bibliographystyle{ACM-Reference-Format}
\balance
\bibliography{2025-www-sp-praise}

\end{document}